\documentclass[sigconf]{acmart}

\copyrightyear{2024}
\acmYear{2024}
\setcopyright{rightsretained}
\acmConference[KDD '24]{Proceedings of the 30th ACM SIGKDD Conference on Knowledge Discovery and Data Mining}{August 25--29, 2024}{Barcelona, Spain}
\acmBooktitle{Proceedings of the 30th ACM SIGKDD Conference on Knowledge Discovery and Data Mining (KDD '24), August 25--29, 2024, Barcelona, Spain}\acmDOI{10.1145/3637528.3671640}
\acmISBN{979-8-4007-0490-1/24/08}

\makeatletter
\gdef\@copyrightpermission{
  \begin{minipage}{0.3\columnwidth}
   \href{https://creativecommons.org/licenses/by/4.0/}{\includegraphics[width=0.90\textwidth]{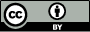}}
  \end{minipage}\hfill
  \begin{minipage}{0.7\columnwidth}
   \href{https://creativecommons.org/licenses/by/4.0/}{This work is licensed under a Creative Commons Attribution International 4.0 License.}
  \end{minipage}
  \vspace{5pt}
}
\makeatother

\begin{document}

\title{Bringing Multimodality to Amazon Visual Search System}

\author{Xinliang Zhu}
\authornote{Contributed equally to this research}
\email{xlzhu@amazon.com}
\orcid{0000-0002-4544-2078}
\affiliation{\institution{Amazon.com}\city{Palo Alto}\state{CA}\country{USA}}

\author{Sheng-Wei Huang}
\authornotemark[1]
\email{shengweh@amazon.com}
\orcid{0000-0002-0244-6335}
\affiliation{\institution{Amazon.com}\city{Palo Alto}\state{CA}\country{USA}}

\author{Han Ding}
\authornotemark[1]
\email{handing@amazon.com}
\orcid{0009-0003-1474-8415}
\affiliation{\institution{Amazon.com}\city{Santa Clara}\state{CA}\country{USA}}

\author{Jinyu Yang}
\email{viyjy@amazon.com}
\orcid{0000-0002-7004-3570}
\affiliation{\institution{Amazon.com}\city{Palo Alto}\state{CA}\country{USA}}

\author{Kelvin Chen}
\email{kelchen@amazon.com}
\orcid{0009-0000-5680-7223}
\affiliation{\institution{Amazon.com}\city{New York}\state{NY}\country{USA}}

\author{Tao Zhou}
\email{taozho@amazon.com}
\orcid{0009-0001-4312-8050}
\affiliation{\institution{Amazon.com}\city{Palo Alto}\state{CA}\country{USA}}

\author{Tal Neiman}
\email{taneiman@amazon.com}
\orcid{0009-0005-0198-240X}
\affiliation{\institution{Amazon.com}\city{New York}\state{NY}\country{USA}}

\author{Ouye Xie}
\email{ouyexie@amazon.com}
\orcid{0009-0002-1950-8501}
\affiliation{\institution{Amazon.com}\city{Seattle}\state{WA}\country{USA}}

\author{Son Tran}
\email{sontran@amazon.com}
\orcid{0009-0001-9206-2916}
\affiliation{\institution{Amazon.com}\city{Palo Alto}\state{CA}\country{USA}}

\author{Benjamin Yao}
\email{benjamy@amazon.com}
\orcid{0009-0005-8622-3540}
\affiliation{\institution{Amazon.com}\city{Seattle}\state{WA}\country{USA}}

\author{Douglas Gray}
\email{douggray@amazon.com}
\orcid{0009-0007-3509-582X}
\affiliation{\institution{Amazon.com}\city{Palo Alto}\state{CA}\country{USA}}

\author{Anuj Bindal}
\email{anbindal@a9.com}
\orcid{0009-0007-2571-9950}
\affiliation{\institution{Amazon.com}\city{Palo Alto}\state{CA}\country{USA}}

\author{Arnab Dhua}
\email{adhua@amazon.com}
\orcid{0009-0007-8233-4301}
\affiliation{\institution{Amazon.com}\city{Palo Alto}\state{CA}\country{USA}}
\renewcommand{\shortauthors}{Xinliang Zhu et al.}

\begin{abstract}
Image to image matching has been well studied in the computer vision community. Previous studies mainly focus on training a deep metric learning model matching visual patterns between the query image and gallery images. In this study, we show that pure image-to-image matching suffers from false positives caused by matching to local visual patterns. To alleviate this issue, we propose to leverage recent advances in vision-language pretraining research. Specifically, we introduce additional image-text alignment losses into deep metric learning, which serve as constraints to the image-to-image matching loss. With additional alignments between the text (e.g., product title) and image pairs, the model can learn concepts from both modalities explicitly, which avoids matching low-level visual features. We progressively develop two variants, a 3-tower and a 4-tower model, where the latter takes one more short text query input. Through extensive experiments, we show that this change leads to a substantial improvement to the image to image matching problem. We further leveraged this model for multimodal search, which takes both image and reformulation text queries to improve search quality. Both offline and online experiments show strong improvements on the main metrics. Specifically, we see 4.95\% relative improvement on image matching click through rate with the 3-tower model and 1.13\% further improvement from the 4-tower model.
\end{abstract}

\begin{CCSXML}
<ccs2012>
<concept>
<concept_id>10002951.10003317.10003338.10010403</concept_id>
<concept_desc>Information systems~Novelty in information retrieval</concept_desc>
<concept_significance>500</concept_significance>
</concept>
<concept>
<concept_id>10002951.10003317.10003338.10003346</concept_id>
<concept_desc>Information systems~Top-k retrieval in databases</concept_desc>
<concept_significance>500</concept_significance>
</concept>
<concept>
<concept_id>10002951.10003317.10003347.10003356</concept_id>
<concept_desc>Information systems~Clustering and classification</concept_desc>
<concept_significance>300</concept_significance>
</concept>
<concept>
<concept_id>10002951.10003317.10003359.10003362</concept_id>
<concept_desc>Information systems~Retrieval effectiveness</concept_desc>
<concept_significance>300</concept_significance>
</concept>
<concept>
<concept_id>10002951.10003317.10003371.10003386.10003387</concept_id>
<concept_desc>Information systems~Image search</concept_desc>
<concept_significance>500</concept_significance>
</concept>
</ccs2012>
\end{CCSXML}

\ccsdesc[500]{Information systems~Novelty in information retrieval}
\ccsdesc[500]{Information systems~Top-k retrieval in databases}
\ccsdesc[300]{Information systems~Clustering and classification}
\ccsdesc[300]{Information systems~Retrieval effectiveness}
\ccsdesc[500]{Information systems~Image search}
\keywords{Image Retrieval, Deep Metric Learning, Vision Language Model, Multimodal Search}



\maketitle

% TODO
% % - Rewrite the story from the perspective of bringing modality to VS
% % - Polish the technical part to reflect the recent progresses in DML, and VL
% % - Polish our contribution part to include TealRanger and Multimodal Search system
% % - Check image sources

\section{Introduction}
\label{sec:intro}

\begin{figure}[t]
\includegraphics[width=\linewidth]{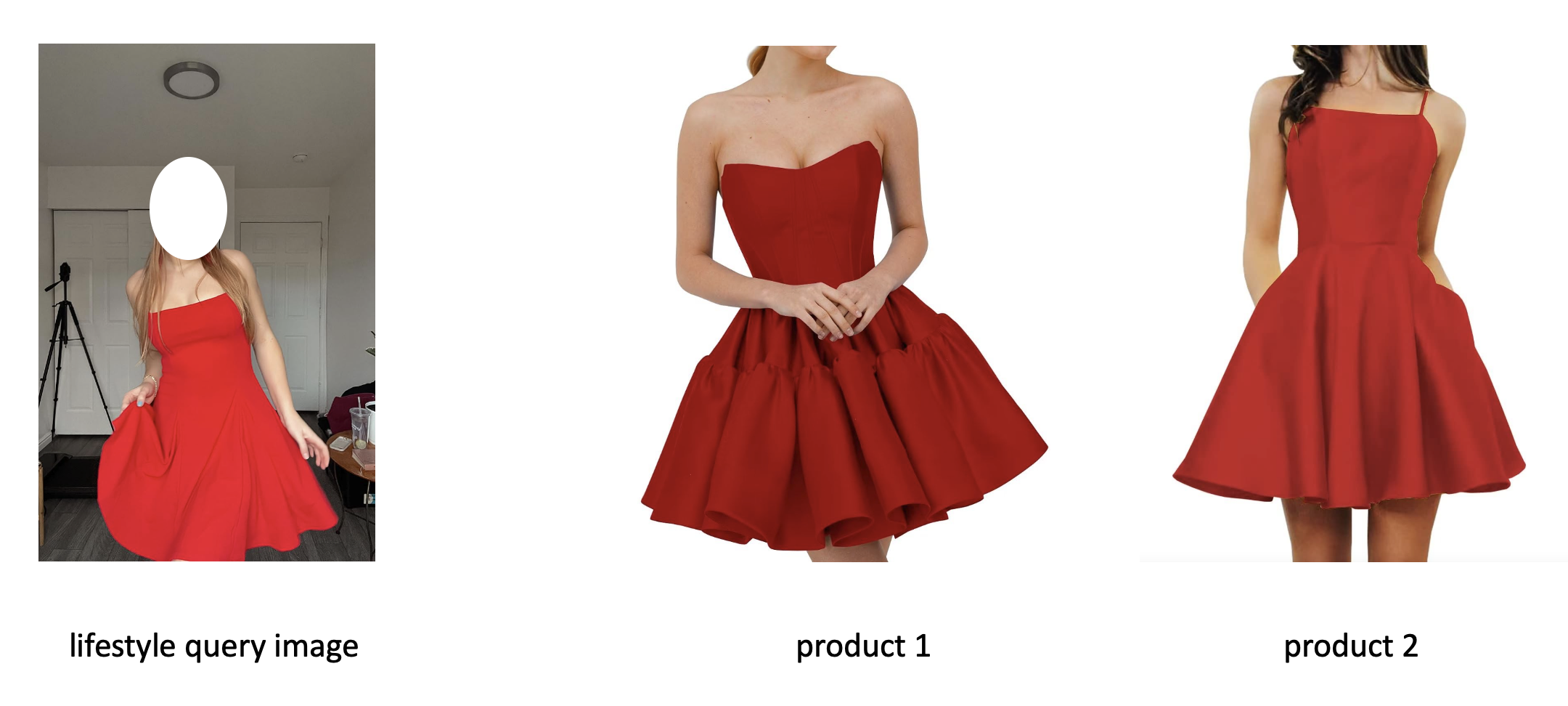} % Check the source of image/ replace it with image.
 \caption{Street-to-shop problem: we use a lifestyle query image (left) to match product images (right) with simple or white background. Note the domain shift between the query and product images.}
 \label{fig:1}
 %\vspace{-3mm}
\end{figure}

Image matching is a well-studied problem in Computer Vision with a wide range of applications. In this work, we focus on the task of matching a visual query to items in a catalog, where queries are typically lifestyle images with cluttered background (i.e., images in the wild) and catalog consists of item images with simple or white backgrounds.  This task, illustrated in Fig.~\ref{fig:1}, and referred to as the street-to-shop problem in previous works~\cite{hadi2015buy,liu2016deepfashion,stl}, is applicable in various real world settings, such as social networks~\cite{bell2020groknet,zhai2019learning}, visual search engines~\cite{zhai2019learning}, and e-commerce websites~\cite{yang2017visual,zhao2019large,stl}. 

When developing a street-to-shop matching system for a customer-facing visual search engine, it is essential to take into account two key aspects: the quality of the training data and the compatibility and scalability of the algorithm. Training data usually consist of image pairs (e.g., a clean product image and a corresponding lifestyle image depicting the same product). Learning algorithms utilize techniques in distance metric learning to embed images in high dimensional space such that street and shop images of the same object are close to each other while images of different objects are far away from each other. Pair based losses such as triplet loss~\cite{schroff2015facenet}, proxy based losses like proxy anchor~\cite{kim2020proxy}, and classification based losses like multi-similarity~\cite{wang2019multi} are examples of commonly used training objectives in deep metric learning (DML). Among them, paired losses are easy to scale up~\cite{manandhar2020dynamically}.

\begin{figure}[t]
\includegraphics[width=\linewidth]{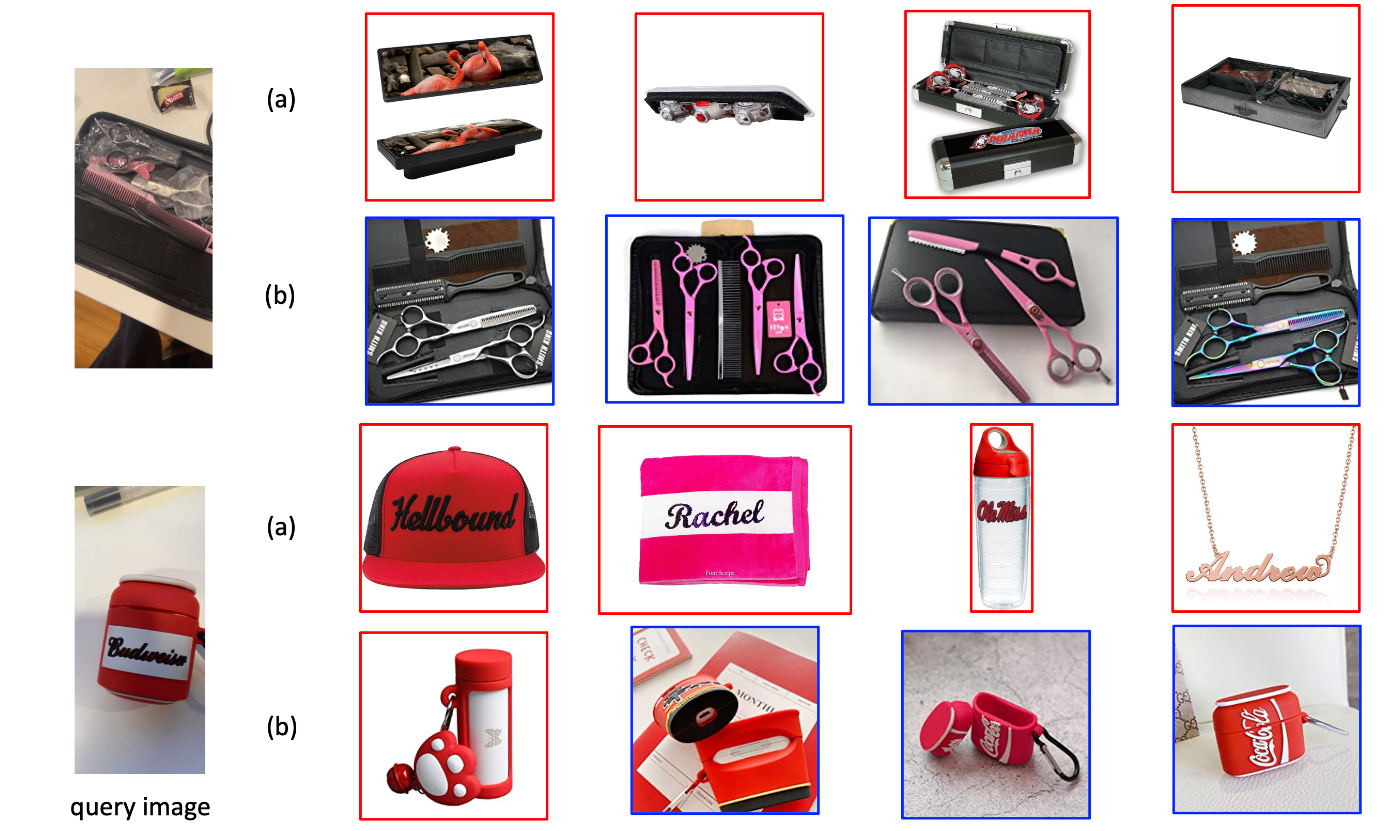}
 \caption{Comparing image match results using traditional pure image-to-image loss versus the proposed method in this paper. Left: query images. Right: retrieved results. Row (a) shows the results from a traditional method and row (b) shows the results from a model trained with the proposed training paradigm. Red boxes mean wrong results. Blue boxes represent exact/similar matches to the query image. In the first query image, a customer tries to find a hair cut set. In the second query image, a customer tries to find an airpod case.}
 \label{fig:2}
 %\vspace{-9mm}
\end{figure}

These algorithms can exactly match certain visual structures well. However, since they are trained on instance-based pairs, they often ignore high level semantic relationships, and as a result, suffer from several practical shortcomings. Fig.~\ref{fig:2} shows the output of one of such algorithm on two example queries. In the first case, the algorithm tends to match irrelevant but dominant features such as background, while placing less weight on the main object. In the second case, matching to local features leads to irrelevant and inconsistent results. The problem is further aggravated when exact matches cannot be found and the outputs are often degraded in an unpredictable and ungraceful manner. Object localization~\cite{redmon2016you}, segmentation~\cite{he2017mask} or weighted salient maps~\cite{wang2018detect} can be used to reduce irrelevant background. However, they often lead to cumbersome models, and increase complexity, or suffer from their own accuracy problems.
One reason for these limitations is that their training data consists of instance-based pairs, which biases the algorithm toward low-level visual feature matching. When the training data has annotated category-level labels, the model might be able to capture higher categorical concepts. There exist datasets with such annotations in the literature such as CUB~\cite{welinder2010caltech} and iNaturalist~\cite{inaturalist}. However, they typically require expensive annotations while having limited coverage, for example, with respect to number of classes. It is challenging to acquire large and comprehensive annotations for categories or attributes especially for complex and diverse man-made objects such as products in an online marketplace. 

In this work we propose to utilize text information associated with images to learn high level concepts for matching. Multimodal signals are available at large scale in public datasets such as LAION ~\cite{LAION}, YFCC100M~\cite{thomee2016yfcc100m}, and CC12M~\cite{changpinyo2021conceptual}. They contain millions to billions of image-text pairs. E-commerce sites and social networks can have up to billions of image-text pairs. Unlike category or attribute labels, the accompanied text is typically unstructured (free text) and complex. Those text can be image captions, web alternative text, text tags or product titles  and descriptions with informative content. In particular, for shop products, the titles often contain key information such as category, various attributes (e.g. material, style, shape, appearance) and fine-grained features. Many of them have corresponding visual meaning which we aim to mine for high level semantic matching. 

\begin{figure}[t]
\includegraphics[width=0.92\linewidth]{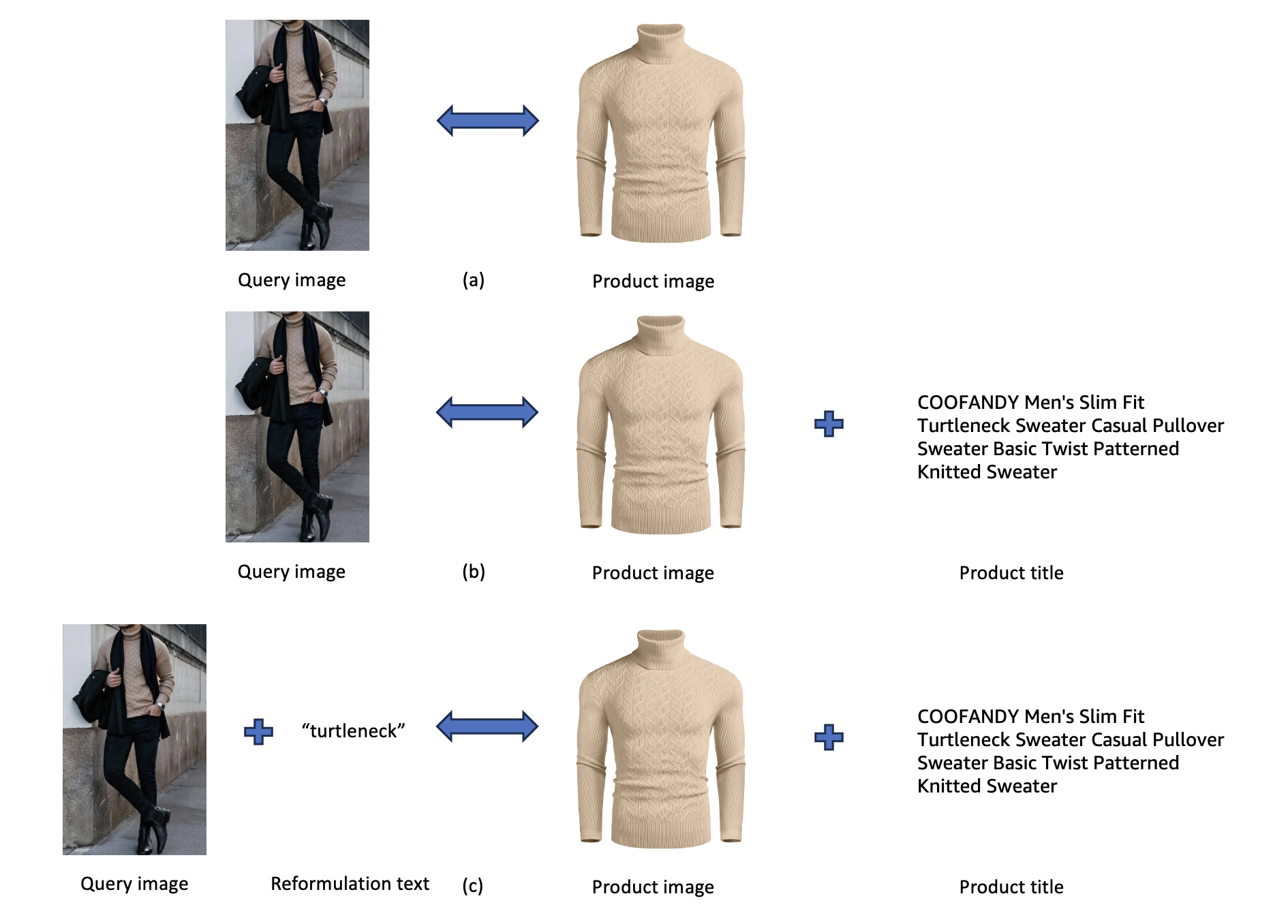}
 \caption{Comparison between existing image match, proposed multimodal image match and multimodal search: (a) existing street-to-shop image matching is query image to product image matching; (b) we propose to use multimodal signals of the products for matching; (c) multimodal search where query image and reformulation text are used to perform the match.}
 \label{fig:dlim_mim_comp}
 %\vspace{-3mm}
\end{figure}

At a high level, our approach is as follows. We align text and image representation into a common embedding space so that matching can be carried out interchangeably across modalities. We design two variants (3-tower and 4-tower models) following the proposed paradigm. In the 3-tower model, we seek to perform alignment in three ways. The first is between query image and catalog image. This is primarily to encourage visual similarity in the output, similar to previous works in image matching~\cite{liu2016deepfashion,hadi2015buy,wang2019multi,proxynca++,dfml,unicom}. The second is between catalog image and its associated text. Its purpose is to unify the embedding space and establish a correspondence between matching visual and language concepts. It is similar to various works in multimodal learning, especially in vision-language pre-training~\cite{clip,jia2021scaling,albef,li2022blip}. The third is between the query image and product text. It is similar to the second alignment, but is a harder task. It helps by reducing the domain gap between lifestyle image, typically with distracting background, and clean text in the catalog. In the 4-tower design, we introduce one more set of alignments between the lifestyle image, catalog image, product title, and a short text query. The short text query is more information dense and cleaner compared to the product title, which further boosts performance. All of these cross modality alignments are carried out using contrastive learning. We name the new method as multimodal image matching (MIM) model. See Fig.~\ref{fig:dlim_mim_comp} for the comparison between traditional image match and proposed MIM and section~\ref{sec:method} for further details. 

At query time, we match the query embedding to a fused signal from catalog items. We chose a simple fusion which averages the embeddings of catalog text and image. From online experiments, we saw 4.95\% relative improvement for the 3-tower model, and a 1.13\% further improvement for the 4-tower model as measured by click through rate. See section~\ref{sec:exps} for further details.

As a by-product of the proposed MIM models, we leverged MIM models for multimodal search, where both query side and catalog side are multimodal. Given the alignment of image and text in MIM models, arithmetic operations can be performed in the latent embedding space when processing both image and text data. Through a combination of offline and online experiments, we demonstrate that employing a straightforward weighted sum of image and text embeddings for query and catalog inputs effectively enhances street-to-shop performance. In practice, users would provide some refinement text (e.g., the type of product they are interested in, its brand, etc.) as the reformulation text into the Multimodal Search system (see Fig.~\ref{fig:dlim_mim_comp} (c) for an example). Although the term ``multimodal search'' shares the same name with the engine proposed in ~\cite{tautkute2019deepstyle}, ours is designed to deal with more scenarios (e.g. reinforcing an attribute, altering an attribute) for more categories (not limited to fashion in ~\cite{tautkute2019deepstyle}).

In summary, our main contributions of this work are:
\begin{enumerate}
    \item  We introduce vision language alignment to the street-to-shop retrieval problem, where both 3-tower and 4-tower models are designed to boost the match performance. To the best of our knowledge, this is the first work in the direction of image retrieval using multimodal signals, especially for the street-to-shop retrieval problem.
    \item We develop a multimodal search system utilizing the proposed MIM models, which stands among the first multimodal search systems in the industry.
    \item Via extensive experiments, we show the effectiveness of our methods from two perspectives: scaling of the model size and scaling of the training dataset size.
\end{enumerate}

The rest of the paper is organized as follows. Section~\ref{sec:related work} reviews the related work. Section~\ref{sec:method} describes our approach in details. Section ~\ref{sec:exps} reports our experimental results. Section~\ref{sec:con} concludes the paper with a discussion about the potential applications of our approach to multimodal search.

% Expand the subsections to include more work and details

\section{Related Work}
\label{sec:related work}
\subsection{Deep Metric Learning} % (Xinliang)
Deep metric learning is vital to modern image matching systems~\cite{zhai2019learning,bell2020groknet,yang2017visual,zhao2019large}. It aims to learn an embedding space, where similar objects are projected close to each other while distinct objects are kept away. The deep embedding models were trained with different types of losses, including paired losses~\cite{hadsell2006dimensionality,schroff2015facenet,hoffer2015deep}, classification-based losses~\cite{movshovitz2017no,zhai2018classification,unicom}, and proxy-anchor based losses~\cite{kim2020proxy,yang2022hierarchical}. To boost the performance of the deep embedding model, some researchers also designed multitask learning methods~\cite{beal2022billion}, and tried to train models with more data~\cite{beal2022billion}. In this paper, we use a type of paired loss (contrastive loss) since it is scalable to a large number of classes (millions) compared to classification-based and proxy-anchor based losses.

\subsection{Vision-language Pretraining} % (Michael)
Recent advancements in vision-language pre-training (VLP) have significantly enhanced our ability to align image and text concepts within a shared latent space. This alignment facilitates a range of applications including visual question answering (VQA), image captioning, cross-modal retrieval, etc. Early works in this direction aim to capture visual-language interaction with a multimodal transformer encoder~\cite{li2019visualbert,li2020oscar,lu2019vilbert, zhang2021vinvl}. These works usually require pre-extracted image and text features and rely on object detectors to align image and text concepts. 

Recent works such as CLIP~\cite{clip} and ALIGN~\cite{jia2021scaling} demonstrated that pre-training dual-encoder with contrastive objectives on web-scale image-text datasets leads to impressive downstream performance (cross-modal retrieval, zero-shot classification etc.). Following the success of CLIP, many works have focused on scaling up training~\cite{cherti2022reproducible} or improving computation efficiency~\cite{gadre2023datacomp,li2023scaling} to further improve downstream performance. 

Research has also been performed on combining dual-encoder architectures with existing learning paradigms. In ALBEF~\cite{albef}, the authors proposed to fuse image and text embeddings with a multimodal encoder and utilize masked language modeling loss. The authors of LiT~\cite{zhai2022lit} combined text encoders with fixed vision encoders that were pre-trained on large-scale image annotation datasets. Authors of Florence~\cite{yuan2021florence} proposed a unified contrastive objective which incorporates both contrastive and cross-entropy losses for dual-encoder training. In DeCLIP~\cite{li2022declip}, the authors combined contrastive loss with SimSiam loss~\cite{chen2020simsiam} and masked language modeling loss in training dual-encoder models, where they observed superior scaling behavior compared to vanilla CLIP model. 

There have also been attempts at tackling various vision-language tasks in one single framework by combining generative objectives with contrastive objectives. In~\cite{tschannen2023image}, the authors studied the scaling property of training a vision encoder with only captioning loss. The trained vision encoder was then combined with a text encoder in LiT~\cite{zhai2022lit} fashion to enable zero-shot evaluations. Authors of CoCA~\cite{yu2022coca} proposed to jointly train a dual-encoder and a multimodal text decoder with both contrastive and captioning losses, which led to state of the art (SOTA) downstream performance. 
% BLIP~\cite{li2022blip} also proposed jointly train a dual-encoder and a decoder with captioning loss, masked language modeling loss and contrastive loss. 
In BEiT3~\cite{wang2022beit3}, the author proposed to train a multi-way transformer with both masked image modeling and masked language modeling.

In this paper, we focus on improving a dual-encoder architecture by training it to align multiple input domains. Specifically, we provide insights on the effectiveness of cross-aligning street-to-shop data in the multimodal context.

\section{Method}
\label{sec:method}

\begin{figure}[htbp]
\includegraphics[width=0.5\textwidth]{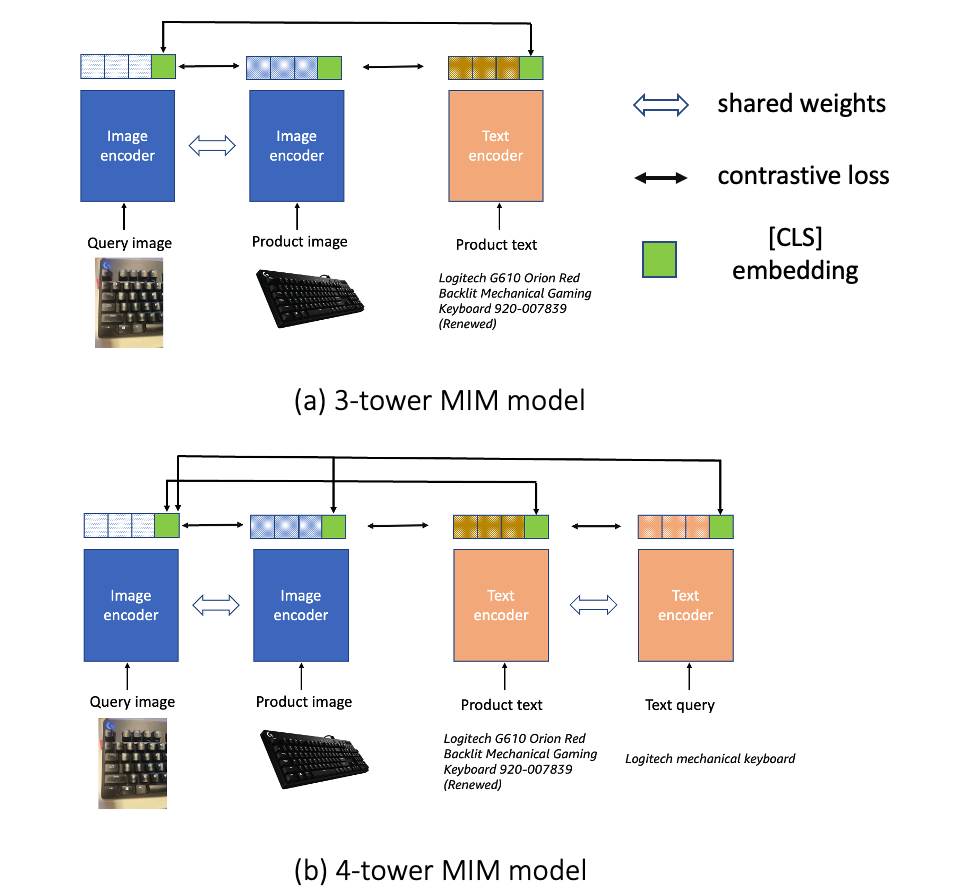}
 \caption{MIM diagram. We develop two variants (3-tower and 4-tower) of the MIM model. In the 3-tower architecture, we have two image encoders and one text encoder. The two image encoders are for query image and product image respectively, and the text encoder is for processing product text. The weights between two image encoders are shared. In the 4-tower architecture, we add one extra text encoder to process short text queries. The two text encoder weights are shared in the 4-tower architecture.}
 \label{fig:diagram}
 %\vspace{-8mm}
\end{figure}

Considering the multi-modal and multi-entity nature of Amazon data, we develop two models for the street-to-shop style image match problem. The basic version is a 3-tower model (MIM-3-tower, Fig.~\ref{fig:diagram}) based on CLIP~\cite{clip}, which is trained on the triples of \{query image, catalog image, product text\}. The advanced version is a 4-tower model (MIM-4-tower, Fig.~\ref{fig:diagram}), which adds one more text arm to accommodate a new short query text entity in the training data. Compared to the common uses of vision language models (e.g., for VQA, cross-model retrieval), the two models proposed in this paper are demonstrated for the first time at Amazon scale (billions of products) for street-to-shop style product search.

% - Change the writing to be based on CLIP.
% - Introduce YellowRanger model
% - Introduce TealRanger model
% - Introduce Mutlimodal Search system

% \subsection{Model Architecture}
\subsection{3-tower Architecture} % (Xinliang & Han)
In this section, we explain the MIM-3-tower model in detail. As shown in Fig.~\ref{fig:diagram}, it consists of three encoders: an image encoder $e_q(\cdot)$ for query images, an image encoder $e_c(\cdot)$ for catalog images, and a text encoder $e_t(\cdot)$ for product text. The weights between the two image encoders are shared to save GPU memory during training. Compared to the original CLIP model~\cite{clip}, we add two new contrastive losses to align input pairs of 1) query image and catalog image and 2) query image and product text. As with CLIP, we have multiple choices of the vision encoder size (from ViT-B/16~\cite{dosovitskiy2020image} to ViT-g/14). For the text encoder, we use a 12-layer BERT$_{base}$~\cite{devlin2018bert}. 

We train the MIM-3-tower model with two groups of objectives: image-image contrastive learning (IIC) on the image encoder, and image-text contrastive learning (ITC) on the image and text encoders. IIC is designed for matching the query images and catalog images as shown in Fig.~\ref{fig:diagram}.
Specifically, we adopt InfoNCE loss \cite{oord2018representation} as our contrastive learning objective, which uses categorical cross-entropy loss to identify the positive sample amongst a set of negative samples.
Given a query-catalog image pair $(Q, C)$, we define $s(Q, C) = f_q(e_q(Q))^Tf_c(e_c(C))$, where $f_q(\cdot)$ and $f_c(\cdot)$ are two linear project layers that map image embeddings to low-dimensional space.
%so that the correct query-catalog image pairs have higher similarity scores. $g_q(v_{cls})$ and $g_c(v_{cls})$ are the normalized [CLS] embeddings~\cite{dosovitskiy2020image} from the image encoder. 
For each $(Q, C)$ pair in the training batch, we calculate the softmax-normalized query-to-catalog and catalog-to-query similarity as follows:

\begin{equation}
\begin{aligned}
    p^{q2c}(Q) = \frac{exp(s(Q, C)/ \tau)}{\sum_{m=1}^M{exp(s(Q, \tilde{C}_m)/\tau)}}, \\
    p^{c2q}(C) = \frac{exp(s(C, Q)/ \tau)}{\sum_{m=1}^M{exp(s(C, \tilde{Q}_m)/\tau)}},
\end{aligned}
\end{equation}
where $\tau$ is a learnable temperature parameter, $\tilde{C}_m$ and $\tilde{Q}_m$ are category images and query images gathered from all GPUs, respectively. 
Here, $M=B \times N_g$, where $B$ is the training batch size and $N_g$ is number of GPUs.
The rationale of introducing $\tilde{C}_m$ and $\tilde{Q}_m$ is that the success of contrastive learning heavily depends on the number of negative samples, which cannot be easily achieved by in-batch negatives.
Denoting $y^{q2c}(Q)$ and $y^{c2q}(C)$ the ground-truth one-hot similarity, with 1 assigned to the positive pairs and 0 to the negative pairs. The image-image contrastive loss is:
\begin{equation}
\begin{aligned}
   \mathcal{L}_{iic} = \frac{1}{2} \mathbb{E}_{(Q,C)\sim D}[H(y^{q2c}(Q), p^{q2c}(Q)) + \\ 
                H(y^{c2q}(C), p^{c2q}(C))],
\end{aligned}
\end{equation}
where $H(y, p)$ is the cross-entropy between $y$ and $p$. Similarly, we can construct the ITC losses between the query image and text, and the catalog image and text. We refer to the CLIP~\cite{clip} paper for the details for the ITC loss. The full training objectives of MIM-3-tower is:
\begin{equation}
    \mathcal{L} = \mathcal{L}_{iic} + \mathcal{L}_{itc},
\label{eq:full_loss}
\end{equation}

where $L_{itc}$ represents ITC loss. It is worth noting that the ITC loss is the sum of two parts: query image to product text and catalog image to product text. From equation~\ref{eq:full_loss}, The IIC is actually a way in deep metric learning to train an image embedding model. The ITC is from the vision language pre-training field, which can be treated as a constraint for the image to image loss (i.e. IIC) in this work. The new way of applying IIC makes a big difference to large scale street-to-shop image match as it can significantly reduce irrelevant matches which are returned due to partial visual pattern matches (e.g., zebra pattern pants vs. zebra pattern mouse pads).

\subsection{4-tower Architecture} % (Michael)
Encouraged by the success of 3-tower architecture, we experimented with a 4-tower architecture by adding query text as an additional alignment target, illustrated in Fig.~\ref{fig:diagram}. Query text is usually search strings containing key attributes pertaining to corresponding products.
% that customers typed when looking for desired products.
Compared to text in the Amazon catalog (title, product description, etc), query text strings are shorter in length but have higher information density. Query text strings are also composed of words used by customers and therefore reside in a different language domain than the Amazon catalog.

Our intuition behind training with query text strings was two-fold: 1) bridging the language domain gap between shoppers and the Amazon catalog can enhance multimodal search capability of the model. 2) training with text that contains high information density can guide the model toward capturing more essential concepts.

The 4-tower architecture consists of 4 encoder towers (query image, query text, catalog image and product text). We designed the encoders with the same modality to share weights, similar to the 3-tower architecture. During training, the model was supervised with 6 contrastive losses, including every pair of the 4 input types.

\subsection{Multimodal Search} % (Xinliang)
Bringing ITC loss to the shop-to-street retrieval problem not only reduces defects for image match but also brings a new opportunity to develop a system we refer to as Multimodal Search (MMS). The definition of Multimodal Search is that given a multimodal query (e.g., an image and a reformulation text), the system returns relevant products meeting the combined intention of the multimodal query. Mathematically, it can be represented as $\{Q_{1}, Q_{2}, ..., Q_{m}\} \rightarrow \{P_{1}, P_{2}, ...,P_{n}\}$, where $m$ and $n$ are the total numbers of modalities in the query and product side, and $Q_m$, $P_n$ represent a single modality from the query side and product side respectively.

In this paper, we have two modalities (i.e., image and text) for both the query and catalog sides. To make the MMS system work and make use of the popular approximate nearest neighbor search infrastructure in industry, we need a way to convert both multimodal query and catalog into a dense feature vector. There are multiple ways (e.g. training a dedicated neural network to fuse the embeddings) to do the conversion. However, we find the simple weighted sum of image embedding and text embedding works well in practice since the key features from the image and text are aligned in the latent space during training. The formula is as following:

\begin{equation}
    X_{m} = w * X_{image} + (1- w) * X_{text},
\label{eq:multimodal_fusion}
\end{equation}
where $X_m$ is the fused embedding, $X_{image}$ is the image embedding and $X_{text}$ is the text embedding. We use different weight $w$ for query and catalog embeddings. With MMS, users can refine pure image search results (e.g., want a specific brand product) or correct image search results by providing more hints/details (e.g., color, product type).

\section{Experiments}
\label{sec:exps}
%   In this section, we illustrate the experimental settings first. Then we show the implementation details followed by results and analysis.

 \subsection{Experimental Settings}

 % To remove? (Michael)
 \subsubsection{Training Data} 
We collected 100 million images of 23 million Amazon products in the daily life context (in-the-wild) to create a training dataset for 3-tower models. Each in-the-wild image has corresponding product and product metadata, including catalog image, title, brand, product description, etc. In total 100M triples of in-the-wild images, catalog image and product text were constructed. We refer to this dataset as 100M triples dataset. The dataset is multi-lingual, covering products from 17 countries. For detailed statistics refer to Table \ref{tab:data_train}.

 To create training dataset for 4-tower models, we utilized the subset of 100M triples where query text strings are available. The subset consists of 56 million in-the-wild images and 13 million products. As each product is associated with large amount of query text strings, we expanded this subset by sampling up to 9 query strings per product. With this, we constructed a 400M quadruples dataset consisting of query images, query text, catalog images, and product text. Details can be found in Table \ref{tab:data_train}.

 % We collect 13 million of customer review images (CRI) from 2 million of ASINs, which cover all categories. For each CRI, it has an associated ASIN and ASIN's metadata (i.e. main variant catalog image, title, bullet point, product description). We construct the training triples (CRI, catalog image, text) from those data. The catalog image is main variant catalog image. We exclude other variant images of the ASIN to reduce the noise. The text is a concatenation of product title, bullet points, and product description. <- ACVC version

 % To update. (Michael)
 \subsubsection{Evaluation Data} 
 We designed our evaluation protocol to focus on recall performance of both multimodal retrieval and image to multimodal retrieval. We designed these two tasks to reflect how our model is utilized to serve customers, where multimodal search is powered by multimodal retrieval, and image matching is augmented by image to multimodal retrieval. To that end, we created an evaluation dataset with a query set and a index set. The query set consists of 8090 unique products, 46,096 in-the-wild product query images and a total of 46,096 image-text pairs. Query strings were composed with product attributes excluding fields included in the index. On the index side, we sourced 1M distractor products from the Amazon catalog in addition to that of the query dataset. The index consists of catalog images and product titles, enabling us to replicate how products are retrieved for customers in reality. For details of the dataset please refer to Table \ref{tab:data_eval}.

 \begin{table}[t]
     \centering
     \begin{tabular}{cccc}
         \toprule
         Name & \#Products & \#Query Images & \#Samples \\
         \midrule
         100M triples & 23M & 100M & 100M \\ 
         400M quadruples & 13M & 57M  & 400M \\
         \bottomrule
     \end{tabular}
     \caption{Training dataset details for 3-tower and 4-tower model training.}
     \label{tab:data_train}
     %\vspace{-3mm}
 \end{table}

 \begin{table}[t]
     \centering
     \begin{tabular}{lcc}
          \toprule
          Set Name &  \#Unique Products & \#Image-Text Pairs\\
          \midrule
          Query    &  8090             & 46,096 \\
          Index    &  1,008,090        & 1,008,090 \\
          \bottomrule
     \end{tabular}
     \caption{Statistics of evaluation dataset.}
     \label{tab:data_eval}
     %\vspace{-5mm}
 \end{table}

 \begin{table*}[]
    \centering
    \begin{tabular}{ccccccccc}
        \toprule
        Row & Method Name  & Vision Encoder & Text Encoder & Initialization & Training Set & Recall@1 & Recall@5 & Recall@10 \\
        \midrule
        \multicolumn{9}{c}{Image to Image Retrieval} \\
        \midrule
        % Unicom      & 0.0019 & 0.0031 & 0.0037 \\ \\
        1 & Contrastive   & ViT-B/16 & n/a & random & 100M triples & 0.18 & 0.27 & 0.31 \\        
        2 & 3-Tower Model & ViT-B/16 & BERT(64M) & random CLIP & 100M triples & 0.19 & 0.30 & 0.35 \\
        3 & 3-Tower Model & ViT-B/16 & BERT(64M) & Amazon CLIP & 100M triples & \textbf{0.30} & \textbf{0.45} & \textbf{0.51} \\
        4 & CLIP (OpenAI) & ViT-B/16 & BERT(63M) & n/a & n/a           & 0.06 & 0.10 & 0.12 \\
        5 & Amazon CLIP   & ViT-B/16 & BERT(64M) & n/a & Amazon Catalog & 0.04 & 0.07 & 0.09 \\ 
        \midrule
        \multicolumn{9}{c}{Image to Multimodal Retrieval} \\
        \midrule
        6 & CLIP (OpenAI) & ViT-B/16 & BERT(63M) & n/a & n/a            & 0.07 & 0.13 & 0.15 \\
        7 & Amazon CLIP   & ViT-B/16 & BERT(64M) & n/a & Amazon Catalog & 0.07 & 0.13 & 0.15 \\        
        8 & 3-Tower Model & ViT-B/16 & BERT(64M) & Amazon CLIP & 100M triples & 0.37 & 0.51 & 0.55 \\
        9 & 3-Tower Model & ViT-B/16 & BERT(64M) & Amazon CLIP & 400M quadruples & 0.34 & 0.52 & 0.58 \\
        10 & 4-Tower Model & ViT-B/16 & BERT(64M) & Amazon CLIP & 400M quadruples & \textbf{0.38} & \textbf{0.54} & \textbf{0.58} \\
        \midrule
        11 & Amazon CLIP & ViT-g/14 & BERT(354M) & n/a & Amazon Catalog & 0.15 & 0.24 & 0.27 \\
        12 & 3-Tower Model & ViT-g/14 & BERT(354M) & Amazon CLIP   & 100M triples & 0.49 & 0.65 & 0.70 \\
        % 10 & 3-Tower Model & ViT-g/14 & BERT(354M) & Amazon CLIP  & 400M quadruples & TK & TK & TK \\
        13 & 4-Tower Model & ViT-g/14 & BERT(354M) & Amazon CLIP   & 400M quadruples & 0.49 & 0.67 & 0.72 \\
        14 & 4-Tower Model & ViT-g/14 & BERT(354M) & 3-Tower Model & 400M quadruples & \textbf{0.54} & \textbf{0.74} & \textbf{0.79} \\
        \midrule
        \multicolumn{9}{c}{Multimodal Retrieval (Multimodal Search)} \\
        \midrule
        15 & CLIP (OpenAI) & ViT-B/16 & n/a & n/a & n/a            & 0.09 & 0.15 & 0.18 \\
        16 & Amazon CLIP   & ViT-B/16 & BERT(64M) & n/a & Amazon Catalog & 0.10 & 0.18 & 0.21 \\
        17 & 3-Tower Model & ViT-B/16 & BERT(64M) & Amazon CLIP & 100M triples    & 0.38 & 0.52 & 0.57 \\
        18 & 3-Tower Model & ViT-B/16 & BERT(64M) & Amazon CLIP & 400M quadruples & 0.37 & 0.55 & 0.61 \\
        19 & 4-Tower Model & ViT-B/16 & BERT(64M) & Amazon CLIP & 400M quadruples & \textbf{0.57} & \textbf{0.74} & \textbf{0.79} \\
        \midrule
        20 & Amazon CLIP   & ViT-g/14 & BERT(354M) & n/a & Amazon Catalog     & 0.25 & 0.41 & 0.47 \\
        21 & 3-Tower Model & ViT-g/14 & BERT(354M) & Amazon CLIP & 100M triples & 0.53 & 0.68 & 0.72 \\
        % 20 & 3-Tower Model & ViT-g/14 & BERT(354M) & Amazon CLIP & 400M quadruples & TK & TK & TK \\
        22 & 4-Tower Model & ViT-g/14 & BERT(354M) & Amazon CLIP & 400M quadruples & 0.61 & 0.78 & 0.82 \\
        23 & 4-Tower Model & ViT-g/14 & BERT(354M) & 3-Tower Model & 400M quadruples & \textbf{0.64} & \textbf{0.82} & \textbf{0.86} \\        
        \bottomrule
    \end{tabular}
    \caption{Benchmarking of different retrieval sub-tasks. For image to image retrieval, leveraging vision-language pre-training improved recall at all positions (row 2). Positive scaling behavior was observed of the 3-tower and the 4-tower model w.r.t. model size, where utilizing ViT-g/14 and BERT(354M) significantly improves recall performance. Results also indicate that 4-tower model outperforms competing methods on both image to multimodal retrieval and multimodal retrieval. Initializing the 4-tower model with 3-tower model weights (pre-trained on 100M triples) led to the best performing model (row 11, 20).}
    \label{tab:result_benchmark_all}
    %\vspace{-3mm}
 \end{table*}
 
 % To update. (Kelvin and Han) (current draft by Michael)
 \subsubsection{Comparison Methods} For both evaluation tasks (i.e., multimodal retrieval and image-multimodal retrieval), the indices were set up as multimodal, with each entry represented by the average of catalog image and product title embeddings. 
 
For multimodal retrieval, queries were defined as weighted sum of query image and query text embeddings. For image to multimodal retrieval, on the other hand, query image embeddings were utilized as the query vectors. During evaluation, image and text embeddings were first extracted for both query and index sets. The catalog image and product title embeddings were then utilized to construct a K-nearest neighbor index. With the index ready, we iterated through the query set to retrieve top-10 products from the index and measured recall at 1st, 5th and 10th position. Such iteration is performed multiple times in a grid-search fashion to find the best image-text weight pair for the model. After the grid search, the best recall performance of multimodal retrieval (non-zero text-weight) and performance of image to multimodal retrieval (zero text-weight) were recorded.

 % To update. (Han and Xinliang) (current draft by Michael
 \subsubsection{Evaluation Metrics} We benchmarked our methods by evaluating recall performance at 1st, 5th and 10th position. We define correctness as retrieving the ground-truth product of the corresponding query image-text pair.
 
 \subsection{Training Procedure} % (We might need to skip this part or hide the data details)
 
 % We generally follow the settings in the ALBEF paper~\cite{albef} that use a 123.7M parameters BERT$_{base}$ model and a 85.8M parameters ViT-B/16 model. The MIM model was trained for 45 epochs using a batch size of 320 on 8 NVIDIA A100 GPUs. The other settings are the same with the ALBEF model. <- ACVC version
 
 \subsubsection{3-tower Model Training}
 % We utilized 128 NVIDIA A100 Tensor Core GPUs in training our 3-way aligned CLIP model. The CLIP model consists of a ViT-g/14 image encoder (1.2B parameters) and a BERT model with (354M parameters). We set the batch size to be 32k and learning rate to be 1e-4. We pre-trained our model on large-scale Amazon catalog dataset with 2B image-text pairs from scratch before fine-tuning with 3-way alignment loss. In the fine-tuning stage, the model was trained for a total of ~470k iterations, effectively observing 1,540M triples. <- From Michael

 We trained our 3-tower model with 128 NVIDIA A100 Tensor Core GPUs, with a total batch size of 32,768. We set learning rate as 5e-4 with 600 steps of warm up and set weight decay as 0.2. We used a 1B ViT-g/14 as our vision encoder and a 354M BERT as our text encoder, which were pretrained on 2B Amazon catalog image and title pairs. We trained the model with 16 epochs on the 100M dataset which is 1.6B samples seen in total. Images are randomly cropped and resized to $224^2$ without keeping the aspect ratio and text are padded to 77 tokens.
 
 % Michael
 \subsubsection{4-tower Model Training}
 We followed the same training setup as the 3-tower model training in terms of number of GPUs, batch size and warm up process. Considering that this is a fine-tuning process, we reduced the learning rate to 7.5e-5 when training with 4-tower models. We trained the model for ~240k iterations, effectively observing 800M quadruples. During 4-tower model training, we padded the images to square and resized to width of 224 for keeping image aspect ratios. We also increased the maximum number of text tokens to 128. The best model was obtained by fine-tuning the best 3-tower model with 4-tower losses. 

 \subsubsection{Computational Cost}
We see 73ms and 21ms (at 90 percentile) inference latency for extracting features with the image encoder and text encoder respectively on a G5.xlarge instance from AWS.
 
 % \subsection{Results}

 % Michael
 % \subsubsection{Method Benchmarking}
 \subsection{Method Benchmarking}
 To measure the effectiveness of our model, we compared our 3-tower and 4-tower alignment training scheme with public metric learning paradigms, including methods that operate on both uni-modal and multimodal input sources. For comparison with uni-modal models, we compared 3-tower model training with a ViT trained with pure Contrastive Loss~\cite{contrastive_chopra_2005}. For VL pre-training, we compared our method with the publicly available CLIP model and the CLIP model trained from scratch on Amazon catalog data. In the interest of fairness, we compared different methods with the same image backbones. 

 % \subsubsection{Comparing 3-Way and 4-Way Alignment}
 \subsection{Comparing 3-tower and 4-tower Architectrue}
\begin{figure}[t]
    \begin{tabular}{@{}c@{}c}
         \includegraphics[width=0.49\linewidth]{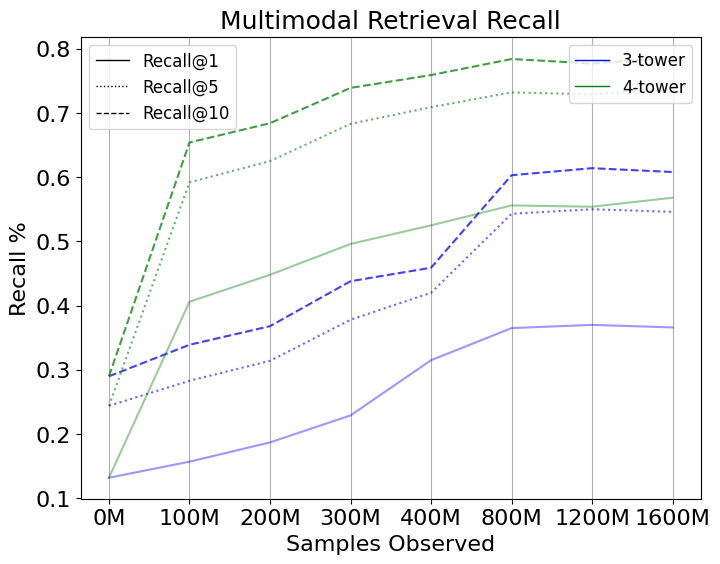} & \includegraphics[width=0.49\linewidth]{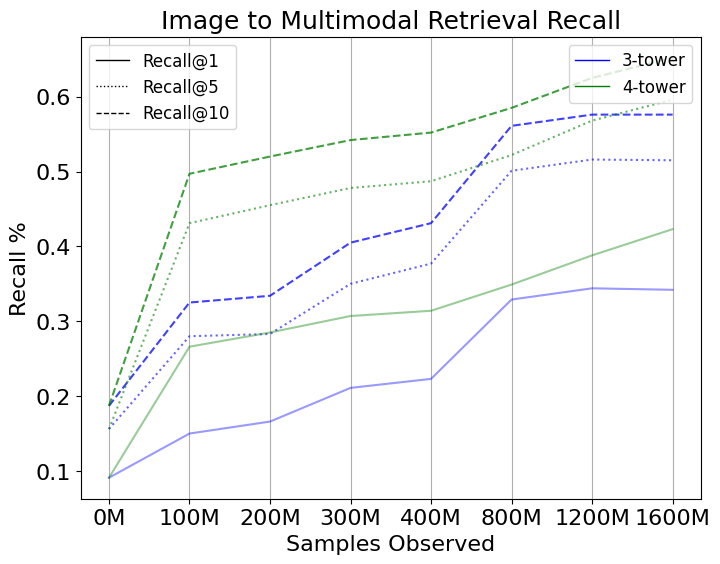} \\
    \end{tabular}
     % % \centering
     % \begin{subfigure}[t]{0.48\linewidth}
     % \centering
     % \includegraphics[width=\linewidth]{fig/towers_compute_impact/mm_scale_tower.png}
     % \end{subfigure}
     % \begin{subfigure}[t]{0.48\linewidth}
     % \centering
     % \includegraphics[width=\linewidth]{fig/towers_compute_impact/im_scale_tower.png}
     % \end{subfigure}     
     \caption{Comparing 3-tower and 4-tower model scaling behavior w.r.t. compute.}
     \label{fig:tower_scaling_behavior}
     %\vspace{-3mm}
 \end{figure}

 With the vision encoder set to ViT-B/16, text encoder set to BERT with 64M parameters and training dataset set to 400M quadruples, the 4-tower model outperforms the 3-tower model in image to multimodal retrieval (see Table~\ref{tab:result_benchmark_all}, row 9 and 10) and multimodal retrieval (Table~\ref{tab:result_benchmark_all}, row 18 and 19). We also observed improved scaling behavior with the 4-tower model with regards to the amount of compute, see figure~\ref{fig:tower_scaling_behavior}. With the same vision backbone, a 4-tower model trained with 400M quadruples still outperforms a 3-tower model trained with 100M triples on both image to multimodal retrieval (Table~\ref{tab:result_benchmark_all}, row 8 and 10) and multimodal retrieval (Table~\ref{tab:result_benchmark_all}, row 17 and 19). This demonstrates the effectiveness of the 4-tower training strategy considering the fact that 400M quadruples dataset contains only 56\% of products and query images of the 100M triples dataset.

 % We also verified this finding with a scaled model version where we used ViT-g/14 as vision encoder and BERT with 354M parameters as text encoder. Under this setup we found the two models to perform on-par on image to multmodal retrieval (Table~\ref{tab:result_benchmark_all}, row 12 and 13) and the 4-tower model outperforming the 3-tower model on multimodal retrieval (Table~\ref{tab:result_benchmark_all}, row 22 and 23). 
We also compared 3-tower and 4-tower model training by using ViT-g/14 as vision encoder and BERT with 354M parameters as text encoder. Compared to a 3-tower model trained on 100M triples, the 4-tower model trained with 400M quadruples achieved superior image to multimodal retrieval performance (Table~\ref{tab:result_benchmark_all}, row 12, 13) despite being trained on a smaller dataset. The 4-tower model also outperformed the 3-tower model in multimodal retrieval (Table~\ref{tab:result_benchmark_all}, row 21, 22).
  
To fully leverage the increased query image and product diversity in 100M triples dataset, we obtained our best 4-tower model by fine-tuning a 100M triples pre-trained 3-tower model with 4-tower training setup. As shown in Table~\ref{tab:result_benchmark_all}, row 14 and 23, the fine-tuned model outperforms every other method in both retrieval evaluation tasks.
 % \subsubsection{Image only versus multimodal Contrastive Learning}

\begin{figure}[h]
    \includegraphics[width=0.49\textwidth]{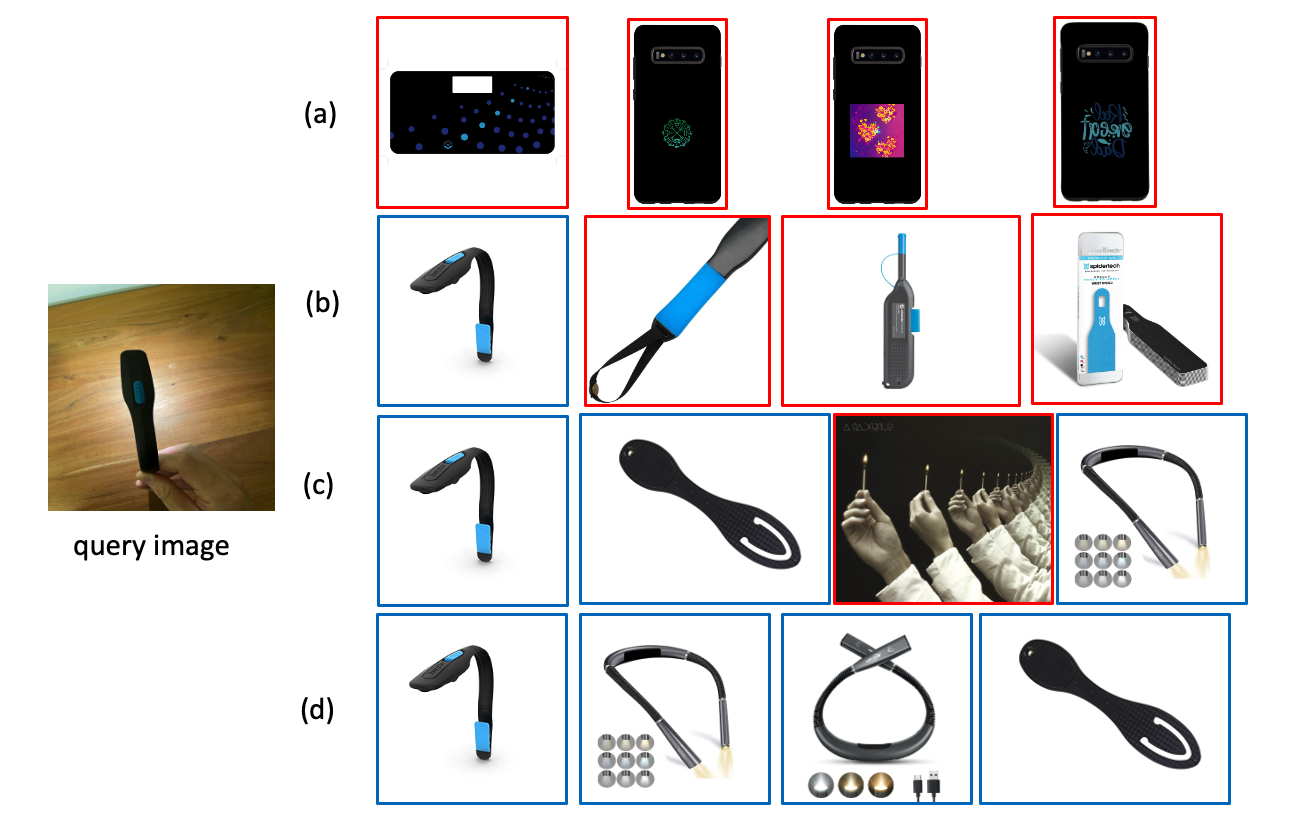}
    \caption{Comparing street-to-shop retrieval from 4 representative models: (a) - (d) show results from Row 1, 2, 12, 13 respectively. Red boxes mean wrong results. Blue boxes represent exact/similar matches to the query image. The model trained with pure image-to-image match loss tends to match partial visual patterns while models trained with additional vision-text alignment loss can find the same or similar products from the same product category. The larger the model, the better the performance. We use our evaluation index to do the comparison, containing less than 1 percent of products compared to our online index, which makes exact matches hard.}
    \label{fig:4-model-comparison}
 %\vspace{-3mm}
\end{figure}
 
 \subsection{Image Only versus Multi-modal Contrastive Learning} 
 In order to understand the impact of incorporating text data in contrastive training, we conducted fine-tuning on the ViT-B/16 CLIP model. This fine-tuning involved using an image-only contrastive loss for one set and the 3-tower image-text contrastive losses for the other. The results, presented in Table~\ref{tab:result_benchmark_all}, reveal performance improvements when 3-tower training is applied.
 % text data is included in the contrastive loss and utilized in index construction. 
Specifically, rows 1 and 2 of the table show that fine-tuning the ViT-B/16-based CLIP model with text data results in increases of 1\% in Recall@1, 3\% in Recall@5, and 4\% in Recall@10.

 \subsection{Impact of Multimodal Index}
 Our experiments show that incorporating text data in contrastive training improves model performance. This led to the hypothesis that integrating text information in the index would similarly enhance image retrieval performance. To test this hypothesis, we conducted an evaluation comparing recall metrics using an image-only index versus a multimodal index. This evaluation was performed on both VL (VL) models and multimodal models. The results, as depicted in Table~\ref{tab:result_benchmark_all}, show notable improvements when text data is added during index construction. Specifically, for the out-of-the-box CLIP model (row 4, 6) and Amazon-CLIP model (row 5, 7), the addition of text data to the index resulted in increases of 1.2\%, 2.5\%, and 3.1\%, and 3.4\%, 5.2\%, and 5.9\% in recall@1, recall@5, and recall@10, respectively. Furthermore, for 3-tower model, the inclusion of text in index building showed even more improvements (row 3, 8) of 7\% recall@1, 6\% recall@5 and 4\% recall@10. Fig.~\ref{fig:4-model-comparison} shows one example of the comparison.

 % \begin{table}[]
 %    \centering
 %    \begin{tabular}{lccc}
 %        \toprule
 %        Index Type   & Recall@1 & Recall@5 & Recall@10 \\
 %        \midrule
 %        \multicolumn{4}{c}{Image to Multimodal Retrieval} \\
 %        \midrule
 %        Image Only   & 0.41 & 0.59 & 0.66 \\
 %        % Text Only    & 0.47 & 0.64 & 0.69 \\
 %        Image + Text & 0.49 & 0.65 & 0.69 \\
 %        \midrule
 %        \multicolumn{4}{c}{Multimodal Retrieval} \\
 %        \midrule
 %        Image Only   & 0.38 & 0.56 & 0.61 \\
 %        % Text Only    & 0.50 & 0.65 & 0.70  \\
 %        Multimodal  & 0.53 & 0.68 & 0.72 \\
 %        \bottomrule
 %    \end{tabular}
 %    \caption{Impact of multimodal index}
 %    \label{tab:results_mm_index_impact}
 % \end{table}

\subsection{Online A/B Testing Results} 
To understand how our models affect customer experience, we performed online A/B testing with our models powering image match and multimodal search functionalities. We considered click-through rate (CTR) as the main criterion, based on the assumption that customers are more likely to click on high-quality retrieved products. We anticipated that improvements in retrieval quality due to new models will therefore be reflected in CTR trends.
 
 As we developed the models progressively, we first conducted online testing for the 3-tower model. In this experiment, we focused on CTR improvement for image matching and monitored how customers interact with the new multimodal search functionality. We observed 4.95\% relative image match CTR improvement, which demonstrated the 3-tower model to be effective in improving image matching quality. 
 We also performed online testing for the 4-tower model, where we observed a further 1.13\% image match CTR improvement and 1.35\% multimodal search CTR improvement. Although the 4-tower model has less image match CTR impact compared to the 3-tower model, we believe this is because the 3-tower model had already significantly improved image matching result quality.

\begin{figure}[h]
    \begin{tabular}{@{}c@{}c}
        \includegraphics[width=0.49\linewidth]{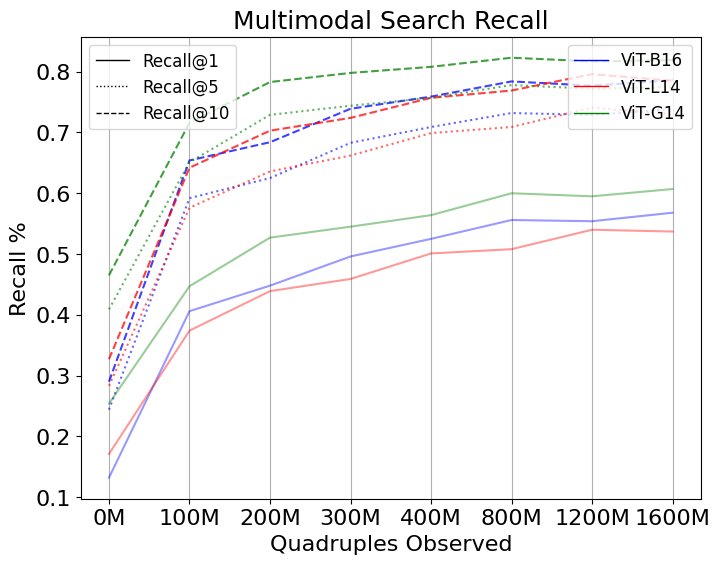} & \includegraphics[width=0.49\linewidth]{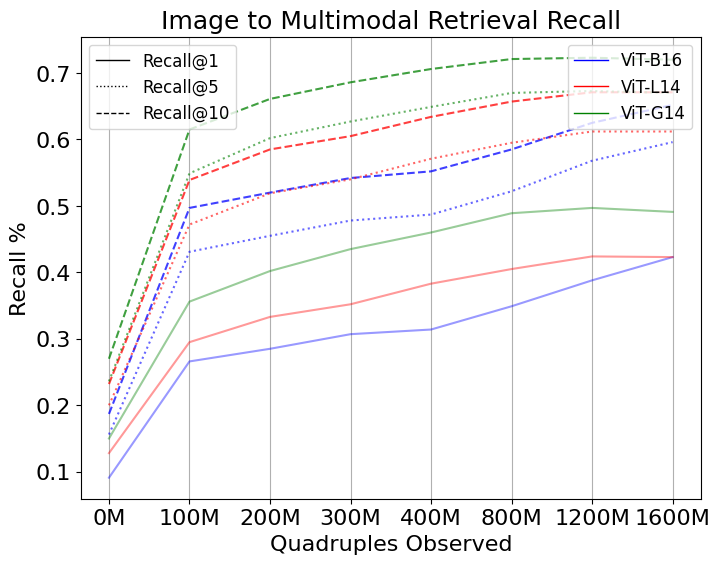} \\
        \includegraphics[width=0.49\linewidth]{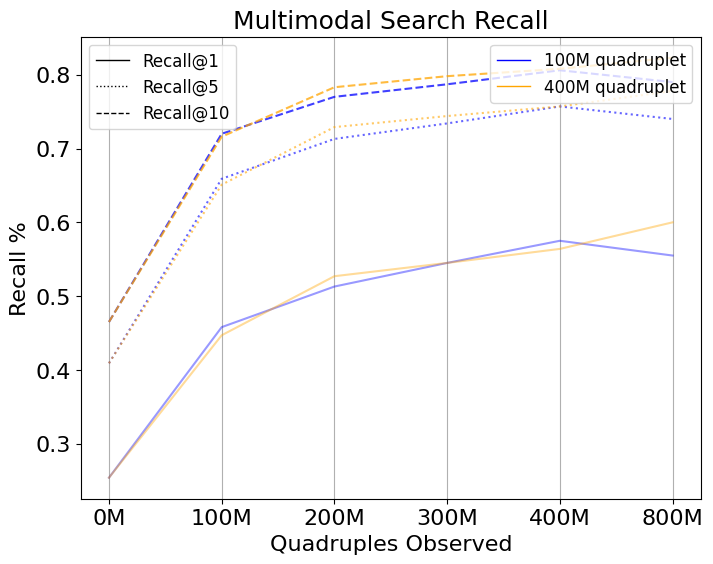} & \includegraphics[width=0.49\linewidth]{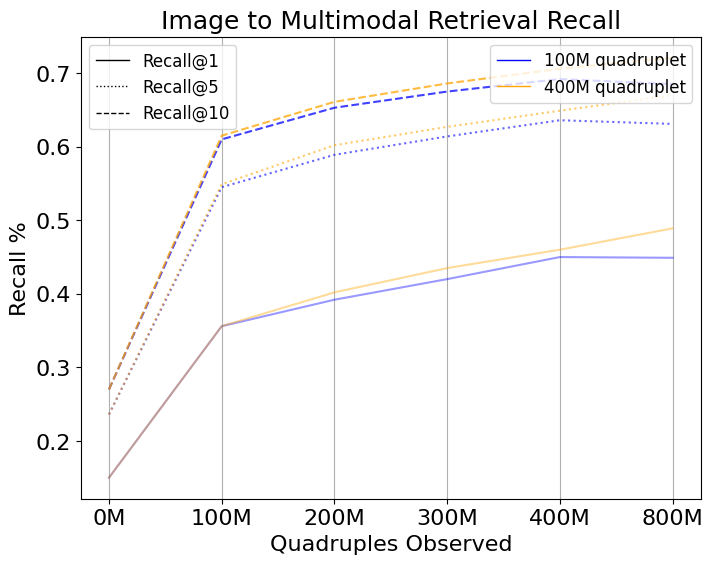} \\
    \end{tabular}
    \caption{Model scaling behavior with different model sizes and dataset sizes. Top two images: Scaling pattern of 4-tower model with different encoder sizes. Bottom two images: Scaling pattern of 4-tower model with different training dataset sizes.}
    \label{fig:ablation_model_size_impact}
%\vspace{-3mm}
\end{figure}

 \subsection{Ablation Study} % (Michael)
 We performed additional experiments to understand how different scaling factors affects model learning (i.e., model size, dataset size). These studies were based on 4-tower architecture, pre-trained with Amazon Catalog data. All experiments were trained with 128 NVIDIA A100 TensorCore GPUs, batch size of 32k and learning rate of 7.5e-5. We utilized a 400M quadruples dataset in these experiments, following the evaluation protocol described in section 4.1.2.
 
 \subsubsection{Impact of Model Size}
 To observe the impact of model size, we trained a 4-tower architecture at 3 different scales, illustrated in Table~\ref{tab:ablation_model_size_impact}. Each model was trained for 4 full epochs, effectively observing 1600M quadruples. Our experiment results indicated that the ViT-g/14 combined with BERT (354M) achieved the best performance on both image to multimodal retrieval and multimodal retrieval. 

 \begin{table}[h]
     \centering
     \begin{tabular}{ccccc}
          \toprule
          Vision Enc & Text Enc & Recall@1 & Recall@5 & Recall@10 \\
          \midrule
          \multicolumn{5}{c}{Image to Multi-modal Retrieval} \\
          \midrule
          ViT-B/16 & BERT(64M)  & 0.42 & 0.60 & 0.65 \\
          ViT-L/14 & BERT(123M) & 0.42 & 0.62 & 0.67 \\
          ViT-g/14 & BERT(354M) & \textbf{0.49} & \textbf{0.67} & \textbf{0.72} \\
          \midrule
          \multicolumn{5}{c}{Multi-modal Retrieval} \\
          \midrule
          ViT-B/16 & BERT(64M)  & 0.57 & 0.74 & 0.78 \\
          ViT-L/14 & BERT(123M) & 0.54 & 0.73 & 0.79 \\
          ViT-g/14 & BERT(354M) & \textbf{0.61} & \textbf{0.78} & \textbf{0.82} \\
          \bottomrule
     \end{tabular}
     \caption{Impact of model scale has on 4-tower model training. Using ViT-g/14 and BERT (354M) led to the best performance, indicating the scalability of our method.}
     \label{tab:ablation_model_size_impact}
     %\vspace{-5mm}
 \end{table}

 \begin{table}[h]
     \centering
     \begin{tabular}{cccc}
          \toprule
          Dataset Size & Recall@1 & Recall@5 & Recall@10 \\
          \midrule
          \multicolumn{4}{c}{Image to Multi-modal Retrieval} \\
          \midrule
          100M quadruples & 0.45 & 0.63 & 0.69 \\
          400M quadruples & \textbf{0.49} & \textbf{0.67} & \textbf{72} \\
          \midrule
          \multicolumn{4}{c}{Multi-modal Retrieval} \\
          \midrule
          100M quadruples & 0.55 & 0.74 & 0.79 \\
          400M quadruples & \textbf{0.6} & \textbf{0.78} & \textbf{0.82} \\
          \bottomrule
     \end{tabular}
     \caption{Impact of dataset size on 4-tower model training. All experiments use ViT-g/14 as image encoder and BERT (354M) as text encoder. Training with 400M quadruples led to better retrieval performance across the board.}
     \label{tab:ablation_dataset_size_impact}
     %\vspace{-5mm}
 \end{table}
% \vspace{-10mm}

 \subsubsection{Impact of Dataset Size}
 We also measured how dataset size impacts model performance. To that end, we subsampled a 400M quadruples dataset and created a 100M quadruples subset. We trained a 4-tower model with the 100M quadruples subset for 8 epochs and compared it with the same model trained with the full 400M quadruples dataset. In this experiment we used ViT-g/14 as the image encoder and BERT (354M) as text encoder. We found that in early iterations the two models scaled similarity in terms of retrieval performance. However, with more compute, the model trained with 400M quadruples dataset clearly outperforms its counterpart as illustrated in Fig.~\ref{fig:ablation_model_size_impact}.
% Update to reflect the two contributions (VL for DML, MMS).
 \section{Conclusion}
 \label{sec:con}
In this paper, we propose a new algorithm named MIM for Amazon scale street-to-shop retrieval problem. MIM can improve performance on the street-to-shop retrieval problem by leveraging text information for visual semantic matching. We also develop a multimodal search service to further improve search quality and give users the flexibility to refine search results. Both offline and online experiments verify the effectiveness of MIM models and the multimodal search service. It shows that image-text alignment is beneficial to the street-to-shop problem as it can learn the product concepts well and avoid false positives due to matching low-level visual features.  In the future, we plan to resolve some limitations of the proposed method such as: 1) the vision model may prioritize retrieving approximate matches over exact matches for some cases; 2) in Multimodal Search still underperforms on cases where lexical match is more important than semantic understanding (e.g., size/part numbers).

\bibliographystyle{abbrv}
\balance
\bibliography{egbib}

\end{document}